\documentclass[acmtog,nonacm]{acmart}

\usepackage{booktabs} 

\usepackage[ruled]{algorithm2e} 

\SetAlFnt{\small}
\SetAlCapFnt{\small}
\SetAlCapNameFnt{\small}
\SetAlCapHSkip{0pt}

\acmJournal{TOG}

\usepackage{adjustbox}
\usepackage{wrapfig}
\usepackage{multirow}
\usepackage{import}
\usepackage{tikz}
\usepackage{graphicx}

\begin{document}
\title{Meshtryoshka: Differentiable Rendering of Real-World Scenes via Mesh Rasterization}

\author{David Charatan}
\authornote{Equal contribution; order decided by coin flip.}
\orcid{0000-0002-1223-4475}
\affiliation{
    \institution{Massachusetts Institute of Technology}
    \city{Cambridge}
    \state{Massachusetts}
    \country{United States}
}
\email{charatan@mit.edu}

\author{Daniel Xu}
\authornotemark[1]
\orcid{0009-0003-4982-6671}
\affiliation{
    \institution{Massachusetts Institute of Technology}
    \city{Cambridge}
    \state{Massachusetts}
    \country{United States}
}
\email{danielxu@mit.edu}

\author{Richard Szeliski}
\orcid{0009-0005-5300-5475}
\affiliation{
    \institution{University of Washington}
    \city{Seattle}
    \state{Washington}
    \country{United States}
}
\email{szeliski@cs.washington.edu}

\author{George Kopanas}
\orcid{0009-0002-5829-2192}
\affiliation{
    \institution{Google DeepMind}
    \city{San Francisco}
    \state{California}
    \country{United States}
}
\email{george.kopanas@gmail.com}

\author{Vincent Sitzmann}
\orcid{0000-0002-0107-5704}
\affiliation{
    \institution{Massachusetts Institute of Technology}
    \city{Cambridge}
    \state{Massachusetts}
    \country{United States}
}
\email{sitzmann@mit.edu}

\renewcommand\shortauthors{Charatan, D. and Xu, D. et al}

\begin{abstract}
Differentiable rendering has emerged as a powerful approach for 3D reconstruction and novel view synthesis.
Today, state-of-the-art differentiable rendering methods combine a variety of custom representations of 3D geometry and appearance with specialized renderers.
However, most downstream tasks in computer graphics rely on 3D meshes, which provide superior portability, allow for hardware-accelerated rendering, and are at the core of most computer graphics workflows. 
While prior work has attempted differentiable rendering with mesh representations, these approaches are limited to object-centric scenes and fail to reconstruct large-scale, unbounded scenes.
In this work, we introduce \textit{Meshtryoshka}, a novel mesh differentiable rendering framework that combines an off-the-shelf triangle rasterizer with a 3D representation that consists of nested mesh shells which resemble a matryoshka doll. 
In every forward pass, the mesh shells are extracted anew from a 3D signed distance function via iso-surface extraction, and the opacities for each vertex are computed as a function of signed distance.
Each mesh shell is then rasterized independently, and the final image is created via alpha compositing.
Crucially, mesh vertex positions are updated only indirectly via gradients that flow through the opacity values into the signed distance function, and hence, our method is compatible with off-the-shelf mesh renderers that need not be differentiable with respect to vertex positions.
On object-centric scenes, our method performs competitively with surface-based differentiable rendering techniques. 
Our differentiable mesh rendering method scales to unbounded, real-world 3D scenes, where it yields high-quality novel view synthesis results approaching those of state-of-the-art, non-mesh methods.
Our method suggests that it may be possible to solve the differentiable rendering problem without relying on specialized renderers, only using conventional tools from the computer graphics toolbox.
Project website: \url{https://danielxu9393.github.io/meshtryoshka-website/}.

\end{abstract}

%
%


%
%
\begin{teaserfigure}
\includegraphics[width=\textwidth]{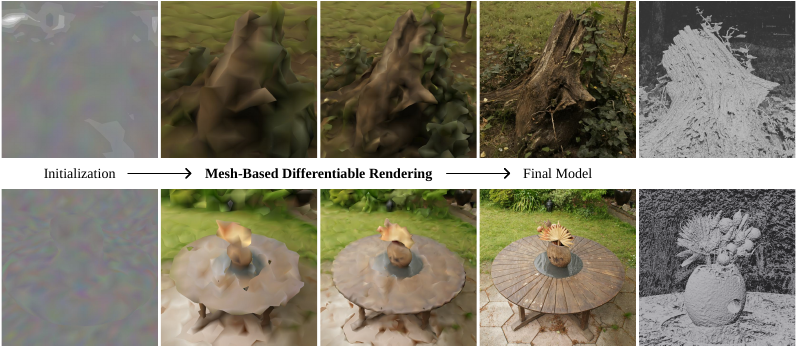}
\caption{
We present \textit{Meshtryoshka}, a mesh-based differentiable rendering method that is capable of reconstructing both object-centric and real-world scenes.
Our method extracts multiple level sets of a signed distance function, individually renders the resulting triangle meshes using a non-differentiable rasterizer, and then alpha-composites the resulting images to produce a rendered view.
Unlike previous mesh-based differentiable renderers, our method is able to reconstruct real-world, unbounded scenes.
}
\label{figure:teaser}
\end{teaserfigure}


\maketitle

\section{Introduction}
Since the dawn of computer graphics, researchers have proposed countless pairings of 3D scene representations and renderers, each offering distinct trade-offs of speed, quality, editability, and simplicity.
Nevertheless, the most widely used 3D representation remains the 3D surface mesh. Meshes are portable, support fast rendering via dedicated hardware implementations, and are essential to many computer graphics workflows, including modeling, animation, and physics simulation.
Yet despite the general consensus surrounding 3D surface meshes, differentiable rendering operates primarily through alternatives to traditional mesh rendering.
In differentiable rendering, we render a 3D scene representation, compare it with ground-truth images to compute a loss, and backpropagate into the representation to reconstruct an underlying scene.
Variants have emerged that differ in their 3D representations, parameterizations, and rendering formulations, each with pros and cons.
However, there is no differentiable renderer that both achieves photorealistic results on real-world scenes and directly optimizes a mesh.

Common wisdom holds that surface meshes are difficult to optimize.
To circumvent this issue, differentiable volume rendering \cite{mildenhall2020nerf} and volume-surface hybrids~\cite{wang2023neus2,yariv2021volume} leverage volumetric density as an easy-to-optimize representation, avoiding local minima related to changing topology and transporting primitives via gradient descent.
Nonetheless, some mesh-based works have achieved impressive reconstruction results by parameterizing surface meshes as the output of iso-surface extraction algorithms and optimizing an underlying signed distance function~\cite{shen2021dmtet,shen2023flexicubes}.
Yet such methods are constrained to simple, object-centric scenes, require foreground-background masks, and fail to reconstruct unbounded, real-world 3D scenes.

In this paper, we introduce a differentiable rendering framework that enables the direct optimization of mesh representations of unbounded, real-world 3D scenes. 
At the core of our method are grids of signed distance values and spherical harmonics that are initially dense, but are dramatically sparsified over the course of optimization.
During each optimization step, we extract a set of nested mesh shells at regularly spaced level sets via Marching Cubes; this inspires the name of our method, \textit{Meshtryoshka}.
We visualize an overview of the way we extract the mesh shells in Fig.~\ref{figure:marching_cubes}.
We then independently rasterize each mesh shell into image space and construct final images by alpha-compositing the shells' colors and SDF-derived opacities.
Finally, we backpropagate the resulting rendering error through the alpha-compositing process.

We stress that in contrast to prior work on differentiable mesh rendering, our approach is compatible with off-the-shelf rasterizers.
\emph{We do not require the renderer to be differentiable with respect to mesh vertex positions}.
Vertex locations are optimized only indirectly by backpropagating into the underlying signed distance field.
This enables novel and portable differentiable pipelines that do not rely on custom CUDA implementations~\cite{kerbl2023gaussiansplatting} or stochastic estimation of gradients~\cite{deliot2024transforming}.

We extensively validate our proposed differentiable mesh rendering framework.
On object-centric scenes, we match prior work on differentiable mesh rendering without the use of foreground masks and are competitive with a recent state-of-the-art differentiable surface rendering method, NeuS2~\cite{wang2023neus2}.
We then demonstrate reconstruction of real-world, unbounded 3D scenes with a differentiable mesh renderer.
Where prior differentiable mesh renderers fail, our method succeeds in reconstructing challenging scenes with high visual quality, approaching the quality of state-of-the-art, non-mesh-based differentiable renderers despite using mesh-based optimization.
Finally, by virtue of our direct optimization of a mesh, the quality achieved in the optimization stage is exactly identical to the quality of the final mesh renderings.

In addition to the main method, we contribute a number of high-performance implementations of classic computer graphics algorithms, such as a sparse Marching Cubes algorithm, a software triangle rasterizer that can render more than 100,000,000 triangles at once, and a novel spatial grid arrangement for sparse Marching Cubes, all of which are essential to scaling mesh rendering to real-world, unbounded scenes.
\section{Related Work}
\begin{figure*}[!t]
    \centering
    \includegraphics[width=\textwidth]{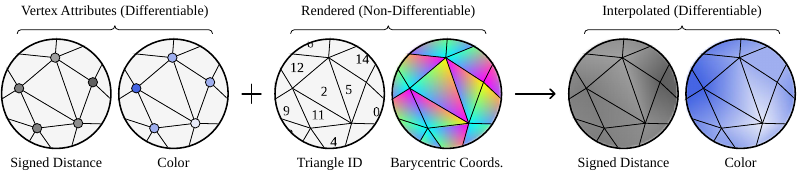}
    \caption{The mesh extraction process (section~\ref{subsection:scene_parameterization_and_mesh_extraction}) yields meshes with per-vertex signed distance values and spherical harmonics coefficients.
    We non-differentiably render these meshes to produce per-pixel triangle ID values and barycentric coordinates.
    After evaluating the spherical harmonics coefficients in the viewing direction to get per-vertex colors, we use differentiable interpolation (akin to a fragment shader) to yield per-pixel signed distance values and colors.
    These are composited using the deferred rendering process described in section ~\ref{subsection:deferred_rendering} to produce a final image.
    }
    \label{figure:sampling}
    \vspace{-1em}
\end{figure*}

\paragraph{Differentiable Isosurface Extraction.}
Classical isosurfacing methods extract a polygonal mesh that approximates the level set of a scalar function.
The most well-known one is Marching Cubes~\allowbreak\cite{lorensen1987marchingcubes}.
Differentiable versions of Marching Cubes~\cite{liao2018deepmarchingcubes, remelli2020meshsdf, guillard2021deepmesh} allow gradients to flow from extracted mesh vertices to the parameters of an underlying surface parameterization.
Other methods~\cite{shen2021dmtet, gao2020deftet} draw upon Marching Tetrahedra~\cite{doi1991marchingtetrahedra} and operate on tetrahedral grids.
FlexiCubes~\cite{shen2023flexicubes} extends Dual Marching Cubes to a multiresolution grid.
%
%
%
%
However, most of these methods are limited by the $O(n^3)$ scaling inherent in regular 3D grids.
To overcome this limitation, we extend Marching Cubes to support sparse data structures, allowing it to scale to high resolutions without the complexity of FlexiCubes' adaptive octree~\cite{shen2023flexicubes}.
%
%
%
Further methods build upon Dual Contouring~\cite{ju2002dualcontouring} and variants of Marching Cubes~\cite{lopes2003improvingtherobustness,chernyaev1995marchingcubes33} to improve isosurface extraction using learned priors.


\paragraph{Differentiable Mesh Rendering.}
Previous work has explored differentiable mesh rendering.
SoftRas~\cite{liu2019softras} introduced an approach where gradients flow through fuzzy triangle edges for image-based 3D reconstruction tasks.
Nvdiffrast \cite{Laine2020diffrast} provided modular primitives for high-performance differentiable rendering, significantly improving computational efficiency and versatility.
Nvdiffrec~\cite{Munkberg_2022_CVPR} jointly optimizes mesh geometry, materials, and lighting from images, achieving high-quality reconstructions.
Shape From Tracing~\cite{goel2020shapefromtracing} optimizes mesh geometry and materials via path tracing.
More recently, DMTet~\cite{shen2021dmtet}, FlexiCubes~\cite{shen2023flexicubes}, and TetWeave~\cite{binninger2025tetweave} introduced methods combining differentiable mesh renderers with differentiable isosurface extraction, enabling robust optimization and detailed geometry reconstruction.
\citet{nicolet2021largesteps} introduces gradient preconditioning methods for faster convergence.
These methods obtain high-quality results on object-centric scenes but generally require a foreground-background mask and cannot reconstruct unbounded, real-world scenes.
They also require the mesh rasterizer to be differentiable with respect to vertex positions.
Our framework matches these methods on object-centric scenes and reconstructs unbounded, real-world scenes.

Concurrent to this work, there has been renewed interest in optimizing triangle meshes directly.
MILo~\cite{guedon2025milo} jointly optimizes 3D Gaussians alongside a triangle mesh extracted using Delaunay triangulation at each training step.
Several methods~\cite{Held20243DConvex, Held2025Triangle,sheng20252dtrianglesplattingdirect,burgdorfer2025radianttrianglesoupsoft} optimize disconnected, semi-transparent triangles that resemble ``triangle soup.''
Triangle Splatting+~\cite{Held2025Triangle+} produces opaque triangles that are compatible with standard rendering engines.
MeshSplatting~\cite{Held2025MeshSplatting} enforces connectivity via Delaunay triangulation.
Our work shares the same goal as MeshSplatting, but relies on sparse differentiable Marching Cubes to achieve connectivity.
Separately, Mesh Splatting~\cite{zhang2026meshsplattingendtoendmultiview} uses a layered rendering that resembles our method's, but uses DMTet~\cite{shen2021dmtet} as its underlying SDF parameterization.

\paragraph{Non-Mesh Differentiable Rendering.}
Differentiable rendering methods optimized for real-world, unbounded 3D reconstruction and photo-realistic novel view synthesis do not rely on meshes during optimization. 
Neural Radiance Fields~\cite{mildenhall2020nerf} instead rely on differentiable volume rendering, generally parameterized via combinations of neural fields, sparse voxel grids, and hash maps~\cite{Sun2022dvgo, mueller2022instant, yu_and_fridovichkeil2021plenoxels, yu2021plenoctrees}.
Surface-based differentiable methods generally also rely on volume rendering, but parameterize density as a function of an underlying surface representation, such as a signed distance field~\cite{wang2021neus, wang2023neus2, yariv2021volume}.
Gaussian Splatting~\cite{kerbl2023gaussiansplatting} parameterizes the geometry as a set of 3D Gaussians, each with view-dependent color and opacity. This parameterization enables real-time rendering and fast optimization by leveraging the rasterization pipeline and approximations of the volume rendering integral.
Contrary to our work, Gaussian Splatting requires a specialized rasterizer, which limits the algorithm's portability, especially during the optimization phase.
To extract surfaces from Gaussian Splatting, recent methods regularize Gaussian primitives towards disks, resembling surfels~\cite{guedon2023sugar,Huang2DGS2024,yu2024gaussian}.
These methods enable extraction of a 3D mesh in a post-processing step via iso-surface extraction.
The highest quality meshes, however, are extracted via more sophisticated ``baking'' processes which further optimize, decimate, or regularize the extracted mesh~\cite{choi2024ltm,yariv2023bakedsdf,Reiser2024SIGGRAPH,chen2022mobilenerf,sharma2024volumetric,wan2023learning,wei2025neumanifold,wang2023adaptive,esposito2024volumetric}.
While these methods enable high-quality novel view synthesis, they extract meshes a posteriori and rely on custom differentiable renderers, non-mesh 3D scene representations, and significant, multi-step processing to ultimately arrive at a 3D mesh.
In contrast, we introduce a novel differentiable rendering framework that achieves competitive novel view synthesis performance by directly optimizing a 3D mesh representation via an off-the-shelf, non-differentiable 3D mesh rasterizer, suggesting that it may be possible to obtain high-quality 3D reconstruction and novel view synthesis while relying on triangle rasterization.

\section{Method}

We transform a standard \emph{non-differentiable} triangle rasterizer into an effective triangle-based differentiable renderer via several key ideas.
First, we parameterize scenes via grids of signed distances\footnote{Because we do not enforce an Eikonal constraint, our learned values are not true signed distances. However, for practical purposes, they can be treated similarly.} and view-dependent colors, allowing us to extract a set of nested mesh shells---a \textit{Meshtryoshka}---from a scene (section~\ref{subsection:scene_parameterization_and_mesh_extraction}).
Second, we render these shells using a two-step pipeline of rasterization and deferred shading, which allows us to build upon an off-the-shelf rasterizer (section~\ref{subsection:deferred_rendering}).
%
Finally, to scale to real-world scenes, we use a mask to only store parameters where needed, efficiently subdivide the grid from coarse to fine, and stretch the grid to efficiently allocate resolution among the center and background of our scene (Section~\ref{subsection:sparsity},~\ref{subsection:real_world_scenes}).
We additionally introduce regularizers to promote sparsity and well-behaved optimization (section~\ref{subsection:regularization}).
We stress that unlike previous approaches~\cite{kerbl2023gaussiansplatting,laine2020nvdiffrast}, ours does not require a custom-built differentiable rasterizer.

\subsection{Scene Parameterization and Mesh Extraction}
\label{subsection:scene_parameterization_and_mesh_extraction}
We represent the scene's geometry with a signed distance function $f_\text{SDF}$.
We do not define this function over the whole domain; instead, we store explicit SDF parameters $\theta_\text{SDF}$ at a set of point locations.
One can imagine these points as a subset of a 3D grid, though in practice they are positioned differently (see section~\ref{subsection:real_world_scenes}).

%
%
During each optimization step, we pass $\theta_\text{SDF}$ to a differentiable Marching Cubes implementation to extract multiple level sets of $f_\text{SDF}$, each forming a separate triangle mesh.
Given an iso-level $\ell$, Marching Cubes creates a triangle vertex wherever neighboring signed distance values lie on opposite sides of $\ell$.
The vertex position is specified via an interpolation weight $t$, where $d_i < \ell < d_j$ are the adjacent distance values:
\begin{equation}
    t = \frac{d_j - \ell}{d_j - d_i}
\end{equation}
We use this interpolation weight to assign a signed distance value $d_\text{vertex}$ to the triangle vertex:
\begin{equation}
    d_\text{vertex} = t d_i + (1 - t) d_j
\end{equation}
While the resulting value of $d_\text{vertex}$ is trivially equal to $\ell$, this equation is used to compute gradients in the backward pass.
We use an additional set of parameters $\theta_\text{color}$, representing view-dependent color, to store spherical harmonics coefficients for each value in $\theta_\text{SDF}$.
We then interpolate values $\mathbf{k}_i$ and $\mathbf{k}_j$ from $\theta_\text{color}$ in the same way as the SDF values to compute spherical harmonics for each triangle vertex:
\begin{equation}
    \mathbf{k}_\text{vertex} = t \mathbf{k}_i + (1 - t) \mathbf{k}_j
\end{equation}
We perform this process on multiple level sets, yielding a collection of nested triangle mesh shells---a \textit{Meshtryoshka}---whose vertices have signed distance values $d_\text{vertex}$ and spherical harmonics coefficients $\mathbf{k}_\text{vertex}$.
\subsection{Non-Differentiable Rasterization \& Deferred Rendering}
\label{subsection:deferred_rendering}
In the previous section, we discussed extracting mesh shells with per-vertex signed distances and spherical harmonics coefficients.
It would be possible to render these shells via ray-tracing, following the formulation from NeuS~\cite{wang2021neus}.
If we followed this approach, we would first compute transmittance values at each ray-triangle intersection:
\begin{equation}
\label{equation:transmittance}
    T_i = \frac{1}{1 + \exp(-d_i)},
\end{equation}
Then, we would convert the transmittance values to alpha values:
\begin{equation}
\label{equation:alpha}
    \alpha_i = \begin{cases}
        1 - T_i & \text{if } i = 1 \\
        \frac{T_{i-1} - T_i}{T_{i-1}} & \text{otherwise}.
    \end{cases}
\end{equation}
Finally, we would alpha-composite the values of $\alpha_i$ together with the corresponding colors (obtained from evaluating the spherical harmonics).
However, this would require us to implement a custom ray-tracer, preventing us from using a non-differentiable, off-the-shelf rasterizer.

We will now discuss an alternative approach, based on deferred rendering, that will allow us to use a non-differentiable triangle rasterizer while still backpropagating gradients into $\theta_\text{SDF}$ and $\theta_\text{color}$.
We observe that because our method creates closely spaced, non-intersecting mesh shells, camera rays will generally intersect the shells sequentially from outermost to innermost.
If the innermost shell's transmittance is chosen to be near zero (i.e., corresponding to a fully opaque surface), then it will fully occlude all further ray-shell intersections, and these further intersections will not contribute to the output image.
Consequently, for each shell, we only need to take the first ray intersection into account.
We thus select the values of $T_i$ as hyperparameters such that the innermost shell's transmittance is near zero and the remaining transmittances are evenly spaced (see Figure~\ref{figure:transmittance}).
Note that by inverting equation~\ref{equation:transmittance}, we can deduce the level sets $\ell_i=d_i$ that correspond to our chosen values of $T_i$.
We use these for the mesh extraction described in section~\ref{subsection:scene_parameterization_and_mesh_extraction}.

Now, observe that given a single shell, an off-the-shelf rasterizer returns exactly what we need---the first ray-shell intersection.
Specifically, at each pixel, it returns the triangle index and the barycentric coordinates (location within the triangle) of the corresponding ray-triangle intersection.
While this operation is non-differentiable, it can be combined with differentiable image-space interpolation to convert per-triangle-vertex values into differentiable per-pixel values.
See Figure~\ref{figure:sampling} for an illustration of this process.
Thus, via differentiable interpolation, we can obtain the same per-pixel signed distances and colors that a custom ray-tracer would produce.

We therefore independently rasterize each shell and interpolate the results to produce per-pixel colors and signed distances.
Next, we use equations \ref{equation:transmittance} and \ref{equation:alpha} to produce per-pixel alpha values.
Finally, we treat the resulting images as transparent layers, alpha-compositing them to produce a final image.
This allows us to backpropagate into $\theta_\text{SDF}$ and $\theta_\text{color}$ despite using a non-differentiable rasterizer.
\begin{figure}[htb]
\centering
\definecolor{cdddddd}{RGB}{221,221,221}
\definecolor{c4565e3}{RGB}{69,101,227}

\def \globalscale {1.000000}
\begin{tikzpicture}[y=1px, x=1px, yscale=\globalscale,xscale=\globalscale, every node/.append style={scale=\globalscale}, inner sep=0pt, outer sep=0pt]
  \path[draw=cdddddd,fill,line cap=round,line width=0.25px] (15.0, 99.5) -- (163.9162, 99.5);

  \path[draw=cdddddd,fill,line cap=round,line width=0.25px] (163.9162, 99.5) -- (163.9162, 14.0);

  \node[text=black,anchor=south] (text1130) at (163.9162, 4.0){$d_0$};

  \node[text=black,anchor=south,cm={ 0.0,1.0,-1.0,0.0,(12.0, -20.5)}] (text6812) at (0.0, 120.0){$T_0$};

  \path[draw=cdddddd,fill,line cap=round,line width=0.25px] (15.0, 80.5) -- (139.3926, 80.5);

  \path[draw=cdddddd,fill,line cap=round,line width=0.25px] (139.3926, 80.5) -- (139.3926, 14.0);

  \node[text=black,anchor=south] (text6414) at (139.3926, 4.0){$d_1$};

  \node[text=black,anchor=south,cm={ 0.0,1.0,-1.0,0.0,(12.0, -39.5)}] (text3982) at (0.0, 120.0){$T_1$};

  \path[draw=cdddddd,fill,line cap=round,line width=0.25px] (15.0, 61.5) -- (124.0, 61.5);

  \path[draw=cdddddd,fill,line cap=round,line width=0.25px] (124.0, 61.5) -- (124.0, 14.0);

  \node[text=black,anchor=south] (text7273) at (124.0, 4.0){$d_2$};

  \node[text=black,anchor=south,cm={ 0.0,1.0,-1.0,0.0,(12.0, -58.5)}] (text8525) at (0.0, 120.0){$T_2$};

  \path[draw=cdddddd,fill,line cap=round,line width=0.25px] (15.0, 42.5) -- (108.6074, 42.5);

  \path[draw=cdddddd,fill,line cap=round,line width=0.25px] (108.6074, 42.5) -- (108.6074, 14.0);

  \node[text=black,anchor=south] (text4879) at (108.6074, 4.0){$d_3$};

  \node[text=black,anchor=south,cm={ 0.0,1.0,-1.0,0.0,(12.0, -77.5)}] (text688) at (0.0, 120.0){$T_3$};

  \path[draw=cdddddd,fill,line cap=round,line width=0.25px] (15.0, 23.5) -- (84.0838, 23.5);

  \path[draw=cdddddd,fill,line cap=round,line width=0.25px] (84.0838, 23.5) -- (84.0838, 14.0);

  \node[text=black,anchor=south] (text8796) at (84.0838, 4.0){$d_4$};

  \node[text=black,anchor=south,cm={ 0.0,1.0,-1.0,0.0,(12.0, -96.5)}] (text5374) at (0.0, 120.0){$T_4$};

  \path[draw=cdddddd,fill,line cap=round,line width=0.25px] (15.0, 16.85) -- (60.8509, 16.85);

  \path[draw=cdddddd,fill,line cap=round,line width=0.25px] (60.8509, 16.85) -- (60.8509, 14.0);

  \node[text=black,anchor=south] (text3045) at (60.8509, 4.0){$d_5$};

  \node[text=black,anchor=south,cm={ 0.0,1.0,-1.0,0.0,(12.0, -103.15)}] (text5233) at (0.0, 120.0){$T_5$};

  \path[draw=c4565e3,line cap=round,line width=1.0px] (15.0, 14.2349) -- (17.202, 14.2651) -- (19.404, 14.2991) -- (21.6061, 14.3375) -- (23.8081, 14.3809) -- (26.0101, 14.4297) -- (28.2121, 14.4848) -- (30.4141, 14.5469) -- (32.6162, 14.617) -- (34.8182, 14.6959) -- (37.0202, 14.7848) -- (39.2222, 14.885) -- (41.4242, 14.9979) -- (43.6263, 15.1249) -- (45.8283, 15.268) -- (48.0303, 15.4289) -- (50.2323, 15.6099) -- (52.4343, 15.8134) -- (54.6364, 16.0421) -- (56.8384, 16.2989) -- (59.0404, 16.587) -- (61.2424, 16.9102) -- (63.4444, 17.2723) -- (65.6465, 17.6777) -- (67.8485, 18.131) -- (70.0505, 18.6373) -- (72.2525, 19.2022) -- (74.4545, 19.8314) -- (76.6566, 20.5312) -- (78.8586, 21.3081) -- (81.0606, 22.1689) -- (83.2626, 23.1205) -- (85.4646, 24.17) -- (87.6667, 25.3243) -- (89.8687, 26.5902) -- (92.0707, 27.974) -- (94.2727, 29.4812) -- (96.4747, 31.1168) -- (98.6768, 32.8841) -- (100.8788, 34.7851) -- (103.0808, 36.8202) -- (105.2828, 38.9874) -- (107.4848, 41.2827) -- (109.6869, 43.6993) -- (111.8889, 46.2281) -- (114.0909, 48.8574) -- (116.2929, 51.5727) -- (118.4949, 54.3576) -- (120.697, 57.1937) -- (122.899, 60.061) -- (125.101, 62.939) -- (127.303, 65.8063) -- (129.5051, 68.6424) -- (131.7071, 71.4273) -- (133.9091, 74.1426) -- (136.1111, 76.7719) -- (138.3131, 79.3007) -- (140.5152, 81.7173) -- (142.7172, 84.0126) -- (144.9192, 86.1798) -- (147.1212, 88.2149) -- (149.3232, 90.1159) -- (151.5253, 91.8832) -- (153.7273, 93.5188) -- (155.9293, 95.026) -- (158.1313, 96.4098) -- (160.3333, 97.6757) -- (162.5354, 98.83) -- (164.7374, 99.8795) -- (166.9394, 100.8311) -- (169.1414, 101.6919) -- (171.3434, 102.4688) -- (173.5455, 103.1686) -- (175.7475, 103.7978) -- (177.9495, 104.3627) -- (180.1515, 104.869) -- (182.3535, 105.3223) -- (184.5556, 105.7277) -- (186.7576, 106.0898) -- (188.9596, 106.413) -- (191.1616, 106.7011) -- (193.3636, 106.9579) -- (195.5657, 107.1866) -- (197.7677, 107.3901) -- (199.9697, 107.5711) -- (202.1717, 107.732) -- (204.3737, 107.8751) -- (206.5758, 108.0021) -- (208.7778, 108.115) -- (210.9798, 108.2152) -- (213.1818, 108.3041) -- (215.3838, 108.383) -- (217.5859, 108.4531) -- (219.7879, 108.5152) -- (221.9899, 108.5703) -- (224.1919, 108.6191) -- (226.3939, 108.6625) -- (228.596, 108.7009) -- (230.798, 108.7349) -- (233.0, 108.7651);;

  \path[draw=black,fill,line cap=round,line width=1.0px] (15.0, 14.0) -- (243.0, 14.0);

  \path[draw=black,fill,line cap=round,line width=1.0px] (243.0, 14.0) -- (240.0, 17.0);

  \path[draw=black,fill,line cap=round,line width=1.0px] (243.0, 14.0) -- (240.0, 11.0);

  \path[draw=black,fill,line cap=round,line width=1.0px] (15.0, 14.0) -- (15.0, 119.0);

  \path[draw=black,fill,line cap=round,line width=1.0px] (15.0, 119.0) -- (12.0, 116.0);

  \path[draw=black,fill,line cap=round,line width=1.0px] (15.0, 119.0) -- (18.0, 116.0);

  \node[text=black,anchor=south east] (text3104) at (237.0, 17.0){$\text{Signed Distance}$};

  \node[text=black,anchor=south east,cm={ 0.0,1.0,-1.0,0.0,(24.0, -7.0)}] (text450) at (0.0, 120.0){$\text{Transmittance}$};

\end{tikzpicture}
\caption{
We select the mesh shells' transmittance parameters $T_i$ as hyperparameters.
Then, by inverting equation~\ref{equation:transmittance}, we deduce the corresponding level sets for iso-surface extraction.
The final $T_i$ is chosen to have a near-zero transmittance, as this allows us to approximate NeuS-style volume rendering using a single intersection per shell (see section~\ref{subsection:deferred_rendering}).
}
\label{figure:transmittance}
\vspace{-1em}
\end{figure}
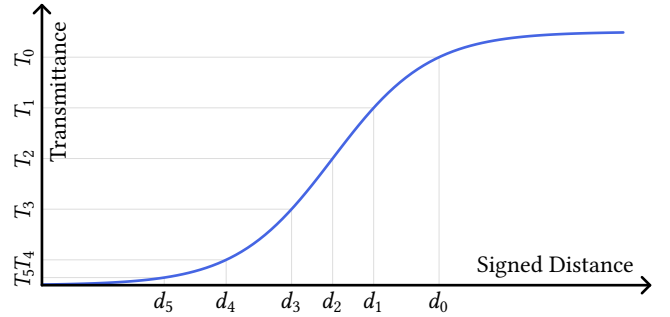
\begin{figure*}[t]
\centering
\includegraphics[width=\textwidth]{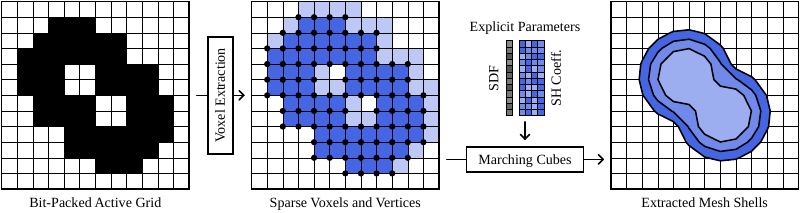}
\caption{
An overview of the mesh extraction process.
Whenever the active grid changes (i.e., after subdivision), the arrays of sparse corner vertices (\textbf{black}), active voxels (\textbf{\textcolor[HTML]{4565E3}{blue}}), and helper voxels (\textbf{\textcolor[HTML]{BDC8F5}{light blue}}) are updated.
These arrays are combined with per-vertex signed distances and spherical harmonics coefficients to produce mesh shells (i.e., a Meshtryoshka) via a sparse, differentiable implementation of Marching Cubes.
}
\label{figure:marching_cubes}
\end{figure*}

\subsection{Sparsity and Coarse-To-Fine Optimization}
\label{subsection:sparsity}
For our method, a high grid resolution is critical to representing fine geometry and textural detail.
However, storing one signed distance value and $3(d + 1)^2$ spherical harmonics coefficients for each grid vertex (where $d$ is the spherical harmonics degree) would require an infeasible amount of memory.
For example, assuming $d = 2$, a dense $1024^3$ grid would require 120 GB of parameters.

Fortunately, voxels that correspond to empty space or solid object interiors do not contribute to the extracted shells, and so it is unnecessary to store parameters at these voxels' corners.
We thus formulate our method around a sparse set of parameters and voxels and adapt Marching Cubes to this setting.
\subsubsection{Voxel Extraction and Representation}
\label{subsubsection:voxel_extraction}
Our sparse scene representation hinges upon a dense 3D grid of binary values which we refer to as the \textit{active grid}.
While this grid also suffers from cubic scaling, its memory consumption is manageable due to the fact that each of its values only requires a single bit.
The active grid indicates which voxels Marching Cubes should run on.
From this grid, we extract flat lists of active vertices and voxels.
A vertex is considered active if it lies on the corner of any active voxel.
Active voxels are represented via flat arrays of metadata: one for the indices of their lower corners (origin, +x, +y, +z), one for upper corners (+xy, +yz, +xz, +xyz), and one for voxel neighbor indices.
The neighbor array stores neighbors in the +x, +y, +z, +xy, +yz, +xz, and +xyz directions for each voxel.
In our implementation, each active voxel can be thought of as ``owning'' the vertex which lies at the corner with the lowest coordinate values in each dimension.
As a result, there exist non-active voxels which ``own'' corners.
We refer to these as \textit{helper voxels} and represent them as additional values at the end of the flat array of lower corners.
Unlike active voxels, helper voxels do not have upper corners or neighbors.
We update the lists of vertices, lower corners, upper corners, and neighbors every time the active grid changes (see section~\ref{subsubsection:subdivision}).
Figure~\ref{figure:marching_cubes} shows an overview.
\subsubsection{Sparse Marching Cubes}
Given the above data structures and a list of per-vertex model parameters, our method follows classical Marching Cubes to extract triangles within each active voxel.
This occurs in two steps: triangle vertex creation and triangle face creation.
The vertex creation step is parallelized over all active voxels and helper voxels; it creates a triangle vertex on a voxel edge whenever the signed distance values at the edge's endpoints are on opposite sides of the level set.
Note that each voxel only creates triangle vertices on edges adjacent to its lowest coordinate, meaning that this step exclusively relies on the lower neighbors and can hence be applied to both active and helper voxels.
On the other hand, the face creation step is parallelized only over active voxels and additionally requires the neighbor array as input.
In this step, we use the neighbor array to deduce the indices of the triangle vertices created by adjacent active and helper voxels.
\subsubsection{Subdivision}
\label{subsubsection:subdivision}
At initialization, our optimization pipeline begins with a fully occupied low-resolution active grid.
At a fixed interval of optimization steps, we prune and subdivide the grid.
To prune the grid, we extract the zero level set of the current scene and mark all voxels that contribute to the zero level set as being active.
We then dilate and upscale the resulting active grid.
After extracting a new set of vertices for the upscaled active grid (see section~\ref{subsubsection:voxel_extraction}), we trilinearly interpolate the previous subdivision level's signed distance and spherical harmonics values.
Because the trilinear interpolation here closely mirrors the one used in Marching Cubes, our subdivision operation only minimally affects the scene's appearance and does not interrupt the optimization process.

\subsection{Non-Cubic Grids for Real-World Scenes}
\label{subsection:real_world_scenes}
\begin{figure}[!t]
\centering
\hspace{-6pt}  \includegraphics[width=\dimexpr\columnwidth+4pt\relax]{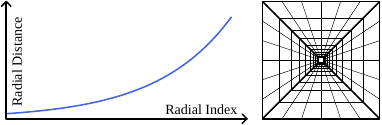}
\caption{
A visualization of the frustum vertices' spatial distribution.
As shown on the left, the vertices' distance from the center increases exponentially.
This allows the model to allocate fewer parameters to far-away parts of the scene.
The exponential vertex placement also ensures that the voxel dimensions (width and height in the 2D illustration on the right) grow in equal proportion to each other.
}
\vspace{-3em}
\label{figure:exponential}
\end{figure}

When fitting our model to object-centric scenes, we use a regular 3D grid in conjunction with the sparse representation described above.
However, real-world scenes cover much greater spatial extents, and so even a sparse regular grid would be too memory-inefficient for our use case.
We therefore divide real-world scenes into foreground and background regions and handle each one separately.
In general, the foreground region is defined as the axis-aligned cube that bounds the cameras, while the background region is defined as all other space.
We use a regular 3D grid to model the foreground region and represent the background region using six truncated frustums (see inset).
\setlength{\columnsep}{8pt}
\begin{wrapfigure}{r}{0.3\linewidth}
\begin{center}
\vspace{-1.25em}
\includegraphics[width=\linewidth]{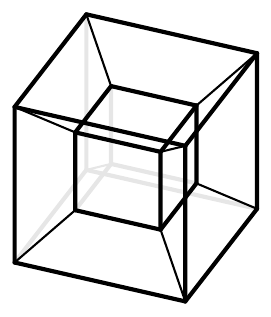}
\vspace{-2em}
\end{center}
\end{wrapfigure}
Each of these frustums contains frustum-shaped voxels that are compatible with the standard Marching Cubes formulation.
To ensure that these voxels are roughly the same size along each dimension, we apply exponential scaling to the voxel corners' distance from the origin.
See Figure~\ref{figure:exponential} for a visualization of this process.
Note that our approach to positioning grid vertices has a similar motivation to the spatial warping and contraction functions from existing NeRF-like methods~\cite{barron2022mipnerf360,neff2021donerf,zhang2020nerfplusplus,reiser2023merf}.
However, as Figure~\ref{figure:exponential} shows, our strategy is particularly suited to our grid-based approach.
\subsection{Regularization}
\label{subsection:regularization}
SDF-based methods often use an Eikonal loss which encourages the SDF gradient magnitude to be 1 everywhere.
However, applying an Eikonal regularizer to explicit grid parameters tends to create zig-zag or checkerboard patterns where each voxel locally satisfies the Eikonal constraint, but the overall signed distance field is unusable.
To circumvent this issue, we instead apply a Laplacian smoothness loss, encouraging each SDF value to match the average of its neighbors.
This promotes smoothness while implicitly regularizing normals and discouraging high-frequency noise.
To prune geometry in unobserved regions, we reward SDF values with high L1 norm, encouraging the values to become strongly positive or negative.
Note that as a result of our choice of regularizers, our signed distance parameters do not form a valid signed distance field---they do not respect the Eikonal constraint.
However, we have presented our method in terms of signed distance fields since the reader is likely to be familiar with them and our method behaves similarly to existing SDF-based methods in practice.
We observe that spherical harmonics cause the model to fall into local minima, where the model can cheat by misrepresenting the geometry to different views.
To limit this effect, we regularize by lowering the learning rate of the non-DC components of the spherical harmonic parameters by 20x and adding an L2 penalty.

\subsection{Implementation}
\label{subsection:implementation}
Our model is primarily implemented using PyTorch~\cite{paszke2019pytorch}, with performance-critical components written using the Slang shader language~\cite{bangaru2023slang}.
We use a Slang-based triangle rasterizer because the preeminent alternative, nvdiffrast~\cite{laine2020nvdiffrast}, only supports up to 16,777,216 triangles.
Our implementation of Marching Cubes is based on the DISO package~\cite{wei2025diso}, but has been ported to Slang and modified to support sparse inputs.
We use 5 mesh shells that correspond to transmittance values of 0.9, 0.5, 0.1, 0.01, and 0.001.
For object-centric scenes, we initialize the signed distance function as a sphere.
For unbounded scenes, we follow Gaussian Splatting~\cite{kerbl2023gaussiansplatting} and initialize $\theta_\text{SDF}$ from an initial point cloud, placing a small sphere around each point.
\section{Results and Evaluation}
We evaluate our model's performance on the NeRF-Synthetic and Mip-NeRF 360 datasets, which we use as representative collections of object-centric and real-world scenes.
We report PSNR, SSIM~\cite{wang2004ssim}, and LPIPS~\cite{zhang2018lpips} on held-out test views. 
\subsection{Baselines}
We compare our method to mesh and non-mesh-based differentiable rendering approaches.
On object-centric scenes, we primarily compare our method to nvdiffrec~\cite{munkberg2022nvdiffrec}, which combines differentiable Dual Marching Cubes with an edge-gradient-based triangle rasterizer.
%
%
Beyond mesh optimization, we compare our method to representative methods from other classes of differentiable renderers: Zip-NeRF for volumetric approaches and 3D Gaussian Splatting for primitive-based ones.
Finally, we compare our approach to Volumetric Surfaces, which bears some similarity to our method because it bakes a NeuS-like model into multiple semi-transparent mesh shells.

\subsection{NeRF-Synthetic}
To evaluate our method on object-centric scenes, we conduct experiments on the NeRF-Synthetic dataset.
Our model is initialized as a sphere at a resolution of 64 and progressively subdivided every 800 training steps, reaching a final resolution of 1024.
We report qualitative results in the second half-page of Figure~\ref{figure:results} and quantitative results in Table~\ref{table:nerf_synthetic}.
Please find further qualitative video results in the Supplemental Material.
Qualitatively, our method succeeds at capturing fine geometric detail.
On the \textsc{lego} scene, our method resolves minute holes in the bulldozer's tracks. 
Our method outperforms nvdiffrec+DMTet, better capturing fine detail, and slightly outperforms nvdiffrec+FlexiCubes in terms of PSNR while performing on par on SSIM and LPIPS.
However, we note that nvdiffrec with both DMTet and FlexiCubes requires visibility masks, while our method does not.
Further, our method obtains performance competitive with a state-of-the-art surface-based, non-mesh rendering method NeuS2~\cite{wang2023neus2}, achieving a PSNR within 0.36 dB. Zip-NeRF, which is not regularized to obtain smooth surfaces, outperforms all other methods.
We further compare against Volumetric Surfaces, which similarly trains a collection of semi-transparent shells, but initializes from the NeuS2 output and does not further deform mesh geometry. For evaluation, we extract their 0-level set surface and assess performance using the same novel view synthesis metrics. We outperform their method by approximately 2 dB.
\begin{table}[!h]
\centering
\begin{tabular*}{\columnwidth}{@{\extracolsep{\fill}}lrrr}
\toprule
Method                 & PSNR $\uparrow$   & SSIM $\uparrow$   & LPIPS $\downarrow$ \\
\midrule
Ours (All Layers)      &             29.36 &             0.939 &              0.077 \\
Ours (Zero Level Set)  &             29.37 &             0.939 &              0.077 \\
nvdiffrec (DMTet)      &             28.80 &             0.938 &              0.078 \\
nvdiffrec (Flexicubes) &             29.22 &             0.940 &              0.076 \\
\midrule
Neus2                  & \underline{29.72} & \underline{0.943} &  \underline{0.068} \\
Zip-NeRF               &    \textbf{33.10} &    \textbf{0.971} &     \textbf{0.031} \\
Volumetric Surfaces    &             27.33 &             0.919 &              0.110 \\
\bottomrule
\end{tabular*}

\caption{
Results on the NeRF Synthetic dataset.
Our method outperforms differentiable mesh rendering methods (top) and approaches the performance of representative volumetric, surface-based, and primitive based methods (bottom).
}

\label{table:nerf_synthetic}

\end{table}

\vspace{-2.5em}
\subsection{Mip-NeRF 360 Reconstruction}
Next, we compare our method to several non-mesh-based differentiable rendering approaches on the Mip-NeRF 360 dataset.
Consistent with observations made in related work~\cite{Reiser2024SIGGRAPH}, our best attempts at reconstructing this dataset with prior mesh-based differentiable rendering method nvdiffrec~\cite{munkberg2022nvdiffrec} with both a sphere-based initialization and a FlexiCubes~\cite{shen2023flexicubes} scene representation failed to produce a mesh that resembled the scene.

\begin{figure*}
\includegraphics[width=\textwidth]{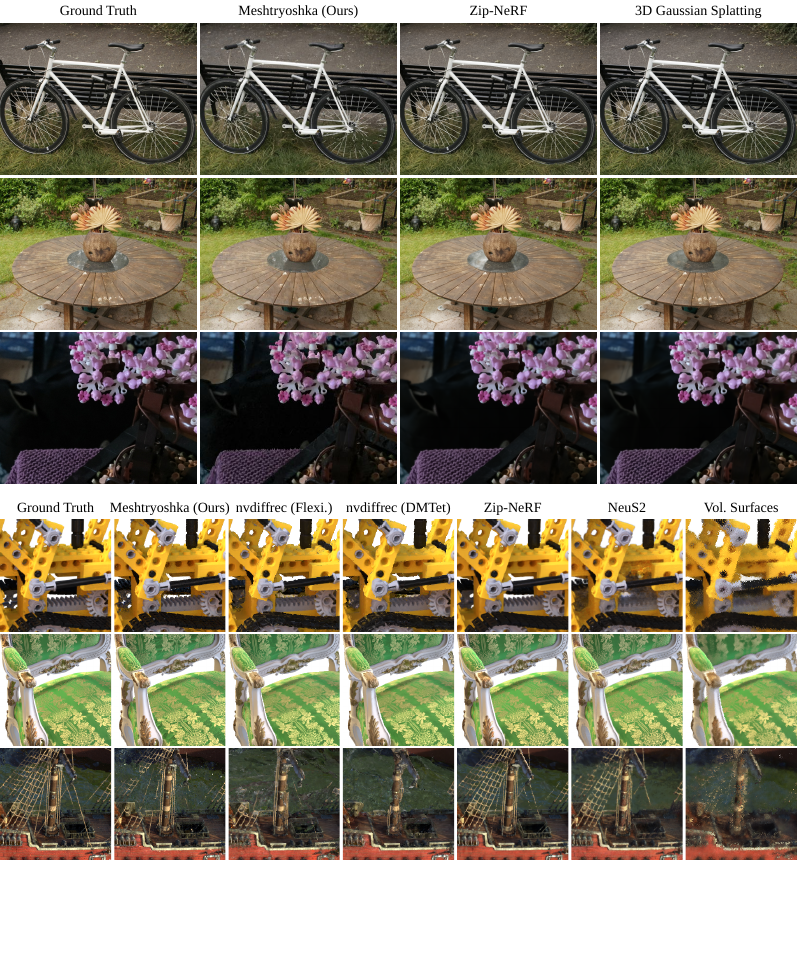}
\vspace{-7.5em}
\caption{
Our method's reconstruction quality on object-centric scenes matches nvdiffrec, an existing mesh-based differentiable rendering framework, and approaches that of non-mesh-based frameworks.
Our method is the only mesh-based differentiable rendering method that is capable of reconstructing real-world scenes, although the visual fidelity it produces is somewhat below that of non-mesh-based methods.
}
\label{figure:results}
\end{figure*}

We show qualitiative results in Figure~\ref{figure:results} and quantitative results in Table~\ref{table:mip_360}. 
Qualitatively, our method obtains crisp and high-quality renderings that capture much of even the high-frequency detail of the scene, such as the spokes of the bike in \textsc{bicycle}, the leaves of trees and bushes in the background of \textsc{garden}, or the fine detail in the leaves of \textsc{bonsai}.
However, our method does not capture quite as many details as Zip-NeRF and displays some high-frequency artifacts around some edges, which are visible with \textsc{bonsai}.
Consistent with this qualitative impression, our quantitative results achieve 24.34 dB, lagging behind Gaussian Splatting by 3 dB, which, in turn, lags Zip-NeRF by about 1.3 dB.
Our method approaches these state-of-the-art methods, validating that mesh-based differentiable rendering is in principle capable of competitive performance and suggesting that custom differentiable renderers may not be critical for competitive real-world differentiable rendering.

\begin{table}[!h]
\centering
\begin{tabular*}{\columnwidth}{@{\extracolsep{\fill}}lrrr}
\toprule
Method                & PSNR $\uparrow$   & SSIM $\uparrow$   & LPIPS $\downarrow$ \\
\midrule
Ours (All Layers)     &             24.34 &             0.686 &              0.367 \\
Ours (Zero Level Set) &             24.30 &             0.684 &              0.369 \\
\midrule
Zip-NeRF              &    \textbf{28.54} &    \textbf{0.828} &     \textbf{0.189} \\
3D Gaussian Splatting & \underline{27.21} & \underline{0.815} &  \underline{0.214} \\
\bottomrule
\end{tabular*}

\caption{
Results on the Mip-NeRF 360 dataset.
Our method is the only differentiable mesh rendering method that is capable of reconstructing real-world scenes (top).
We compare its performance against gold-standard volumetric and primitive based methods (bottom).
}

\label{table:mip_360}

\end{table}

\vspace{-2em}
\subsection{Analysis}
\vspace{-1em}
\label{subsection:analysis}
\begin{figure}[!h]
\centering
\includegraphics[width=\columnwidth]{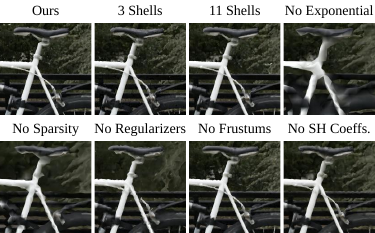}
\caption{
We ablate our method's hyperparameter and design choices.
See section~\ref{subsection:analysis} for an explanation of each ablation.
}
\label{figure:ablations}
\end{figure}
\begin{table}[!h]
\centering

\begin{tabular*}{\columnwidth}{@{\extracolsep{\fill}}lrrr}
\toprule
Method          & PSNR $\uparrow$   & SSIM $\uparrow$   & LPIPS $\downarrow$ \\
\midrule
Ours            &    \textbf{21.87} &    \textbf{0.520} &     \textbf{0.418} \\
3 Shells        &             21.00 &             0.445 &              0.473 \\
11 Shells       & \underline{21.83} & \underline{0.512} &  \underline{0.438} \\
No Exponential  &             19.84 &             0.376 &              0.622 \\
No Sparsity     &             20.53 &             0.405 &              0.616 \\
No Regularizers &             20.55 &             0.410 &              0.569 \\
No Frustums     &             21.56 &             0.489 &              0.471 \\
No SH Coeffs.   &             20.71 &             0.430 &              0.551 \\
\bottomrule
\end{tabular*}

\caption{
We validate our design decisions by ablating several key components of our method.
See section~\ref{subsection:analysis} for descriptions of each ablation.
}

\label{table:ablations}

\end{table}
\vspace{-1em}

To validate our method's design, we ablate several key components on Mip-NeRF 360's \textsc{garden} scene, quantitatively in Table~\ref{table:ablations} and qualitatively in Figure~\ref{figure:ablations}.
In the ``no exponential'' ablation, we distribute our mesh vertices evenly in space rather than using the exponential vertex positioning.
In the ``no sparsity'' ablation, we maintain a constant memory budget and use a lower-resolution dense 3D grid instead of the sparse grid discussed in section~\ref{subsection:sparsity}.
In the ``no regularizers'' ablation, we remove the regularizers discussed in section~\ref{subsection:regularization}.
This ablation demonstrates that our regularizers are essential for well-behaved optimization that yields high-quality results.
In the ``no frustums'' ablation, we use a cubic grid for both the foreground and the background rather than using truncated frustums for the background.
In the ``no SH coeffs.'' ablation, we disable spherical harmonics and use view-independent colors.
These ablations all yield lower quality for the same memory budget.
Finally, we ablate the number of shells used for training in the ``3 shells'' and ``11 shells'' ablations (we generally use 5 shells).
\subsection{Limitations}
\begin{figure}[!h]
\centering
\includegraphics[width=\columnwidth]{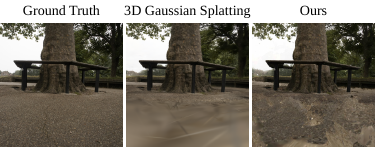}
\caption{
Limitations.
Like Gaussian Splatting, our method struggles to accurately reconstruct low camera angles and fine details in regions that are sparsely covered by the training views.
Gaussian Splatting tends to blur these regions; our method creates undesirable high-frequency artifacts.
}
\label{figure:limitations}
\end{figure}

While our method is able to reconstruct both real-world and object-centric scenes, it has limitations that we hope to address in future work.
First, our method's number of triangles is larger than the number of primitives in existing primitive-based approaches.
This could be addressed by using adaptive iso-surface extraction methods (e.g., based on octrees).
Second, our method's training time---about 4 hours on a single H200 GPU---is slower than that of fast NeRF and 3D Gaussian methods.
Finally, similar to primitive-based Gaussian Splatting methods, our method's optimization dynamics can be less well-behaved than those of NeRF-like methods.
In particular, our method will sometimes create thin ``floaters'' which we show in Figure~\ref{figure:limitations}. Since NeRF was introduced, NeRF-like methods have overcome this challenge with sophisticated regularization techniques. Similar regularization techniques compatible with our representation are left to future work.

\section{Discussion and Conclusion}
We have introduced a novel differentiable mesh rendering framework that leverages a non-differentiable triangle rasterizer to optimize a 3D mesh representation of a scene.
Our work departs from current mainstream approaches to differentiable rendering based on volumetric radiance field representations, offering a fresh theoretical perspective as well as novel algorithmic tools. 
Our method demonstrates that there exists a path towards matching today's impressive differentiable rendering results using only pieces of the conventional computer graphics rendering stack.

\bibliographystyle{ACM-Reference-Format}
\bibliography{references}


\end{document}